\documentclass[letterpaper, 10 pt, conference]{ieeeconf}  % Comment this line out if you need a4paper

\IEEEoverridecommandlockouts                              % This command is only needed if 
\usepackage{cite}
\usepackage{amsmath,amssymb,amsfonts}
\usepackage{algorithmic}
\usepackage{graphicx}
\usepackage{textcomp}
\usepackage{xcolor}
\usepackage{multirow}
\usepackage{adjustbox}
\usepackage{caption}
\usepackage{subcaption}
\usepackage{balance}
\usepackage{float}
\usepackage{url}

\title{\LARGE \bf
Selective Communication for Cooperative Perception in End-to-End Autonomous Driving
}

\author{Hsu-kuang Chiu$^1$ and Stephen F. Smith$^1$% <-this % stops a space
\thanks{$^1$Carnegie Mellon University, Robotics Institute.}%
}

\begin{document}
\balance

\maketitle
\thispagestyle{empty}
\pagestyle{empty}

%%%%%%%%%%%%%%%%%%%%%%%%%%%%%%%%%%%%%%%%%%%%%%%%%%%%%%%%%%%%%%%%%%%%%%%%%%%%%%%%
\begin{abstract}
The reliability of current autonomous driving systems is often jeopardized in situations when the vehicle's
field-of-view is limited by nearby occluding objects. To mitigate this problem, vehicle-to-vehicle communication to share sensor information among multiple autonomous driving vehicles has been proposed. However, to enable timely processing and use of shared sensor data, it is necessary to constrain communication bandwidth, and prior work has done so by restricting the number of other cooperative vehicles and randomly selecting the subset of vehicles to exchange information with from all those that are within communication range. Although simple and cost effective from a communication perspective, this selection approach suffers from its susceptibility to missing those vehicles that possess the perception information most critical to navigation planning. Inspired by recent multi-agent path finding research, we propose a novel selective communication algorithm for cooperative perception to address this shortcoming.  Implemented with a lightweight perception network and a previously developed control network, our algorithm is shown to produce higher success rates than a random selection approach on previously studied safety-critical driving scenario simulations, with minimal additional communication overhead.
\end{abstract}
%%%%%%%%%%%%%%%%%%%%%%%%%%%%%%%%%%%%%%%%%%%%%%%%%%%%%%%%%%%%%%%%%%%%%%%%%%%%%%%%

\section{Introduction}

Safety is the most critical concern when deploying autonomous driving systems in the real world. Although autonomous driving technology has advanced dramatically over the past decade due to such factors as the emergence of deep learning and large-scale driving datasets \cite{geiger2012kitti, caesar2019nuscenes, chang2019argoverse, sun2020waymo, ettinger2021waymo}, current state-of-the-art autonomous driving frameworks still have limitations. One principal challenge on the perception side is that the lidar sensors and cameras are usually attached to the roof of an autonomous driving vehicle, and can be occluded by large nearby objects such as trucks or buses. Such perception uncertainty hinders the ability of  downstream planning and control processes to make safe driving decisions. 
%An example is shown in Figure \ref{fig_motivation}.

One potential solution to this limited field-of-view problem is cooperative perception: using vehicle-to-vehicle
 communication to share perception information with other nearby autonomous driving vehicles. Several prior works \cite{wang2020v2vnet, miller2020cooperative, cui2022coopernaut, xu2022opencood, xu2023v2v4real} have investigated methods for sharing different types of perception information (e.g., lidar point clouds, encoded point features, 3D object detection bounding boxes)  to optimize %perception performance in terms of 
3D object detection accuracy under communication bandwidth constraints. However, higher 3D object detection accuracy does not always produce better driving decisions and safer driving behaviors. For example, cooperative detection of vehicles that are trailing or moving away from a controlled vehicle will likely not improve overall safety much.

%For example, if a cooperative perception algorithm helps a given vehicle to detect another vehicle that is $200$ meters behind it, additional information may not improve overall safety much. 

\begin{figure}[b]
\centering
\includegraphics[width=0.48\textwidth]
{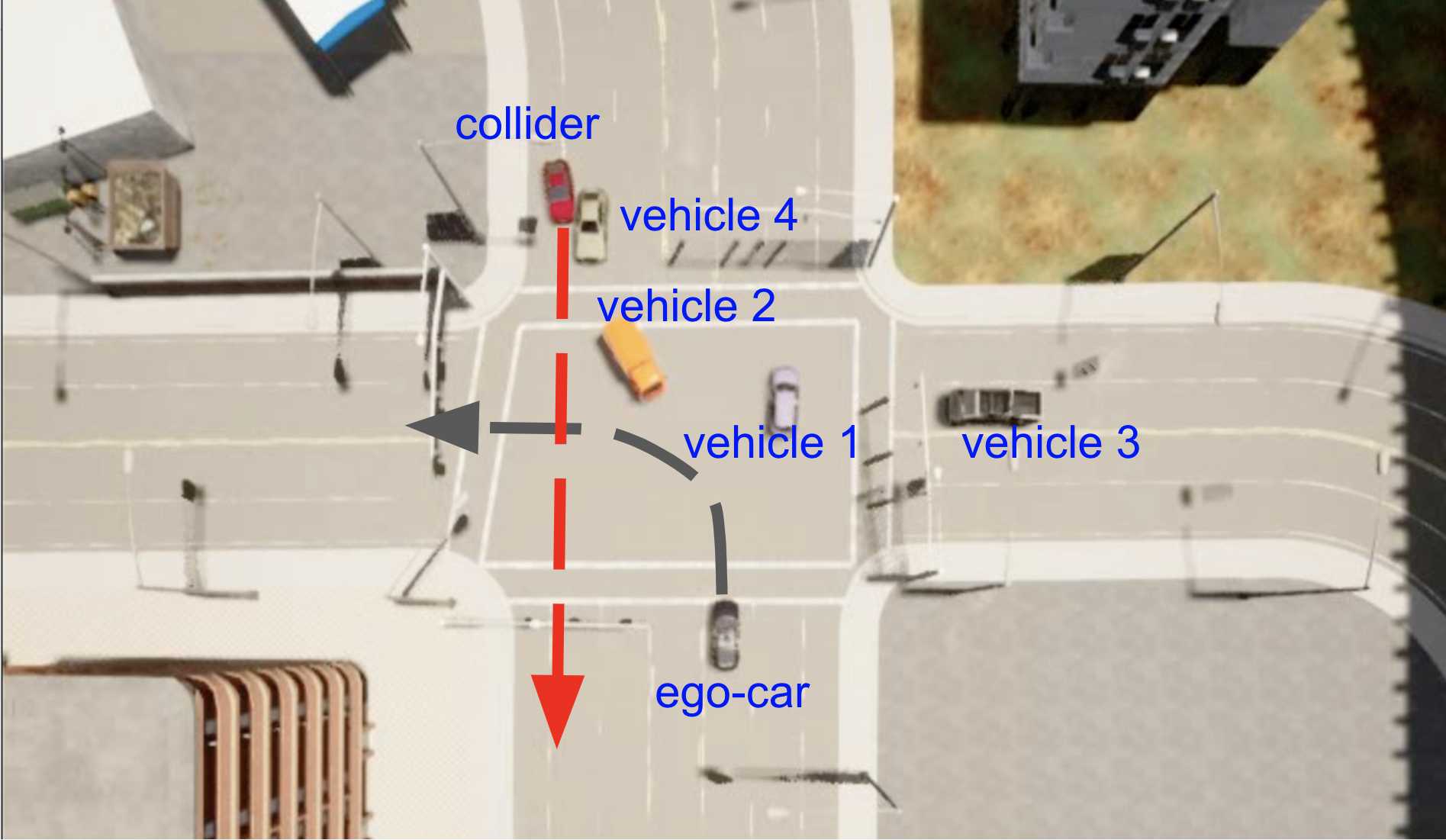}
\caption[]
        {A potential pre-crash driving scenario. The red car crossing the intersection downward is the potential collider and is not communicating with other vehicles. The grey car is the ego-car being controlled by a driving algorithm and plans to turn left. The yellow truck blocks the ego-car's line-of-sight and ability to detect the pre-colliding red car. Vehicles $1 \sim 4$ can all communicate with the ego-car, but only vehicle $4$ can detect the pre-colliding red car given their relative locations. If the ego-car can only select $3$ other vehicles to communicate with due to communication bandwidth constraints, the random selection approach has a $25\%$ chance of causing a collision.
        \vspace{-10pt}} 
        \label{fig_motivation}
\end{figure}

One recent work \cite{cui2022coopernaut} addresses this evaluation issue by proposing an end-to-end autonomous driving neural network with cooperative perception, applying it to a set of safety-critical driving scenarios within the CARLA \cite{dosovitskiy2017carla} autonomous driving system simulator, and measuring the resulting success rates. 
%One recent work \cite{cui2022coopernaut} addresses this evaluation issue by creating safety-critical driving scenarios in the CARLA \cite{dosovitskiy2017carla} autonomous driving systems simulator, and measuring the success rates by simulating their proposed end-to-end autonomous driving neural network based on cooperative perception as input. 
To respect communication bandwidth constraints, their algorithm limits the scope of cooperative perception with the autonomous vehicle being simulated (referred to hereafter as the {\em ego-car}) to a subset of the other vehicles within communication range, and this subset is determined through random selection. 
%randomly chooses the subset of other nearby vehicles from which to obtain perception information, and 
Because of differences in potential utility of the perception information that can be provided by different nearby vehicles, this approach causes high variance in the output control signals generated and the overall driving success rates obtained. Moreover, this random selection approach may omit those nearby vehicles who possess safety-critical perception information relevant to the ego-car's planned trajectory and are thus the most informative, as shown in Fig. \ref{fig_motivation}.  

Inspired by the idea of {\em selective communication} in multi-agent path finding research \cite{ma2022learning}, we propose a novel selective communication algorithm for determining the subset of other vehicles that can best cooperatively improve the perception coverage of the ego-car, given the same constraints on subset 
size. This algorithm is coupled with a lightweight 3D object detector to minimize additional computational cost and communication overhead. We then apply the same driving neural network and evaluation settings as used previously in \cite{cui2022coopernaut} to show that our selective communication algorithm effectively improves success rates for the set of safety-critical driving scenarios previously considered, with only a small increase in communication bandwidth.
%Inspired by the idea of {\em selective communication} in multi-agent path finding research \cite{ma2022learning}, we propose a novel selective communication algorithm and develop a simplified lightweight 3D object detector to let the ego-car determine the subset of other vehicles that can best cooperatively improve its perception coverage, given the same constraints on subset size. Then we apply the same driving neural network and evaluation settings as used previously in \cite{cui2022coopernaut} to show that our selective communication algorithm effectively improves success rates for the set of safety-critical driving scenarios previously considered, with a small increase in communication bandwidth.

\section{Related Work}

\subsection{Autonomous Driving}
State-of-the-art autonomous driving vehicles 
%deployed in urban regions at large scale
\cite{sun2020waymo,ettinger2021waymo} operate mainly by executing a sequence of modular functional components, including detection\cite{yin2020center, dung2020sfa3d, li2020rtm3d}, tracking\cite{kalman1960filter, weng2019ab3dmot, weng2020gnn3dmot, chiu2020probabilistic, chiu2021probabilistic}, prediction\cite{chai2019multipath, ivanovic2019trajectron, salzmann2020trajectronpp}, planning\cite{bronstein2022hierarchical}, and control. This approach provides human-understandable decision-making procedures, but errors may propagate through the systems.
%from earlier components to later ones. 

An alternative approach that has emerged more recently in the research community is end-to-end autonomous driving \cite{bojarski2016end}, which directly produces the final control signals based on sensor input without performing the aforementioned intermediate tasks. Both the algorithm proposed in\cite{cui2022coopernaut} and the more informed variant proposed in this paper are instances of this end-to-end approach. %Our proposed algorithm additionally employs a simplified lightweight 3D object detector to determine which other vehicles possess safety-critical perception information worthy of sharing with the host vehicle.

\subsection{Multi-agent Path Finding}

The concept of cooperative perception by multiple autonomous vehicles has much in common with variants of multiagent path finding (MAPF) problem \cite{li2023intersection, ma2022learning}, and given this similarity, techniques developed to solve these types of MAPF problems can inspire new solutions to cooperative autonomous driving problems. In particular, \cite{ma2022learning} proposes a decentralized {\em selective communication} algorithm for a MAPF problem where each agent only has local partial visibility of the navigation environment. Each agent selects other relevant agents in its field-of-view and uses their local observations to produce better path planning.

%Each vehicle in the aforementioned state-of-the-art autonomous driving systems performs the perception and control tasks individually using its own sensor input and treats other nearby autonomous driving vehicles simply as dynamic obstacles. A better approach is considering information sharing among them to make safer driving decisions. Therefore, techniques from multi-agent path finding research \cite{li2023intersection, ma2022learning} can also inspire new ideas for cooperative autonomous driving problems. For example, \cite{ma2022learning} proposes a decentralized selective communication algorithm for a multi-agent path finding problem where each agent only has local partial observation of the navigation environment. The ego-agent selects other relevant agents in its field-of-view and uses their local observations to produce better path planning.

To be sure, the navigation environment considered in \cite{ma2022learning} is different than the autonomous driving setting of interest in this paper. It is discretized in both space and time, and each agent's field-of-view is simply a 2D square region without considering visual occlusion. Nonetheless, the idea underlying this work has inspired our proposed selective communication algorithm for cooperative perception.

\subsection{Cooperative Perception}
Given an algorithm for determining the set of other vehicles to communicate with,
%Once the agents to communicate with are determined by the selection algorithm, 
another important design consideration is what kind of information or data should be communicated from participating vehicles.
%to effectively improve perception coverage. 
In \cite{arnold2022cooperative} the sharing of lidar point clouds with nearby vehicles is advocated. Other research \cite{miller2020cooperative, saxena2019multiagent} has proposed the sharing of 3D object detection or tracking results. More recent work \cite{zhang2021distributed, xu2022opencood, cui2022coopernaut} has favored sharing more efficient, encoded point features, which achieves a good balance between the performance and communication cost and is adopted in this paper. In our proposed selective communication algorithm, each vehicle first shares relatively sparse center positions of detected objects with the ego-car to determine the subset of vehicles that  possess the most useful perception information. Then, in a second round of communication, those selected vehicles share their relatively dense point features with the ego-car for producing its final control signals.

\section{Method}
In the subsections below, we first formally define the problem of interest in this paper. Next we describe our selective communication algorithm. 
%uses decentralized perception networks to determine the best set of vehicles to communicate with. 
Finally, we summarize the model architectures of the perception neural networks used to assess relative information gain and the control neural networks (drawn from \cite{cui2022coopernaut}) used to derive control signals.

\subsection{Problem Definition}
Given a driving scenario with a static urban environment and a set of other dynamic vehicles (e.g., Fig. \ref{fig_motivation}), we aim to develop a driving algorithm capable of generating control signals that eventually move the ego-car from its starting location to the goal location without colliding with either the environment or other vehicles within a finite time interval. At each time step, the input to the driving algorithm is limited to the ego-car's encoded lidar observation feature vector $O^{ego}$ and the encoded lidar observation feature vectors $(O^{c_1}, O^{c_2}, ...,  O^{c_{N_c}})$ of  $N_c$ selected vehicles. %, as shown in Figure \ref{fig_system_architecture}.

We define the \textbf{selection scope} as :
\begin{align}
    S_s = \{s_1, s_2, ..., s_{N_s}\},
\end{align}
where each $s_i$ represents the index of a vehicle in ego-car's communication range. We define the \textbf{communication scope} as:
\begin{align}
    S_c = \{c_1, c_2, ..., c_{N_c}\},
\end{align}
where each $c_i$ represents an index of a vehicle selected to share their encoded observation information with the ego-car. The size of $S_s$,  $|S_s| = N_s$, and the size of $S_c$, $|S_c| = N_c$, are given as constant parameters. % reflecting communication range and bandwidth constraints.

Our immediate goal is to develop a \textbf{selection algorithm} for determining the best subset $S_c \subset S_s$ of vehicles to communicate with. Once determined, the encoded lidar observation features $(O^{c_1}, O^{c_2}, ...,  O^{c_{N_c}})$ of all vehicles in $S_c$ are sent to the ego-car. We also aim to incorporate a \textbf{driving algorithm} that takes $(O^{ego}, O^{c_1}, O^{c_2}, ...,  O^{c_{N_c}})$ as input and generates the driving control signals: $throttle \in [0, 1]$, $brake \in [0, 1]$, and $steer \in [-1, 1]$ at each time step. %As mentioned above, the overall goal is to produce control signals that can move the ego-car to the goal location within a finite time without collision.

\begin{figure}[ht!]
\centering
\includegraphics[width=0.48\textwidth]{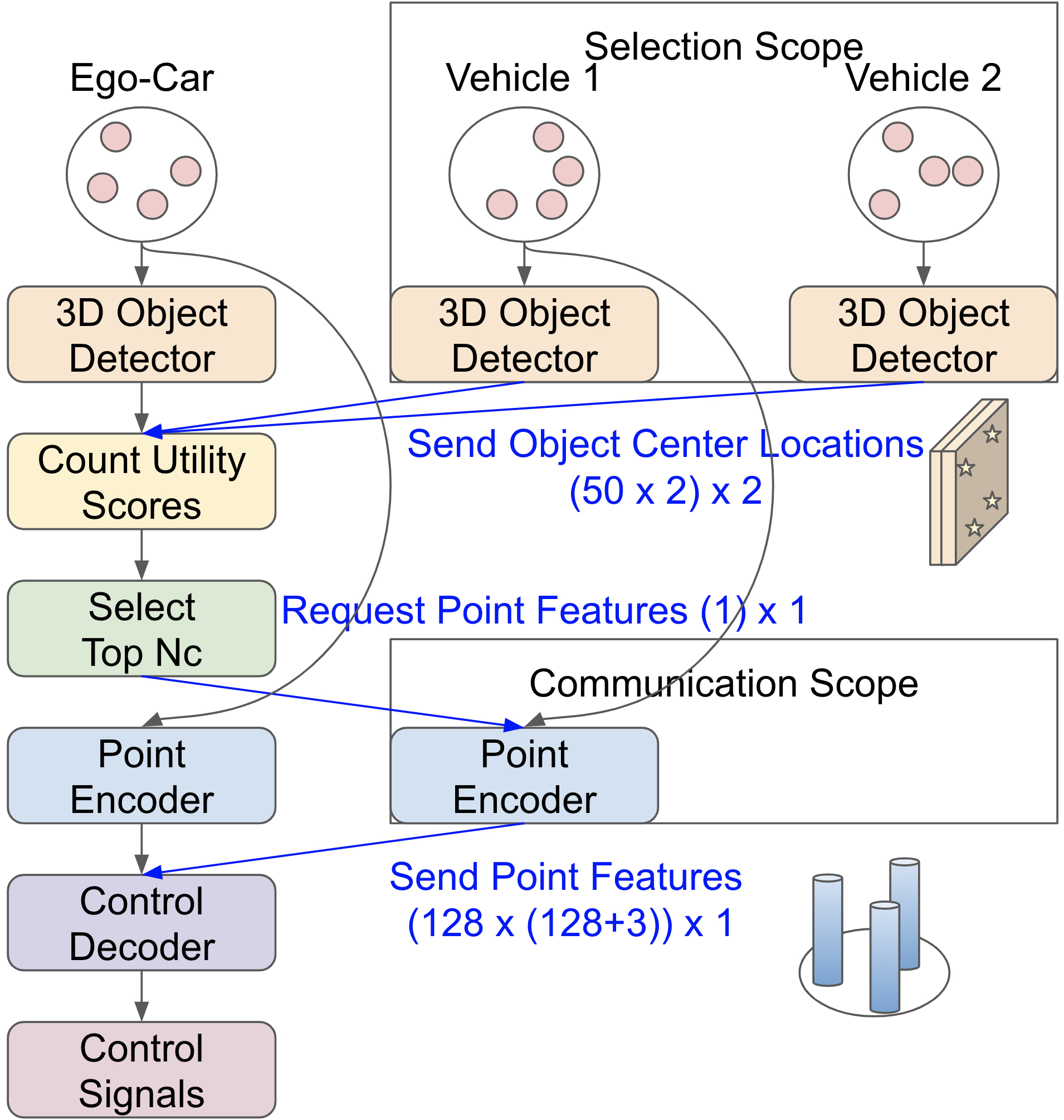}
\caption[]
        {Two-phase selective communication procedure and control signal generator applied to a minimal example (The numbers shown at each communication step indicate how many floating point numbers are transmitted).} %Each vehicle in the selection scope first uses its input lidar point cloud and a 3D object detector to detect objects and sends the center locations of all detected objects to the ego-car. The ego-car then uses this information to estimate the information gain that would be provided by each prospective communicating vehicle by counting the number of detected objects that the ego-car missed, and the subset of vehicles with the highest estimated gains are selected to form the communication scope. The ego-car then requests and receives the encoded point features from all vehicles in the communication scope. Finally, the ego-car aggregates all the features and produces the control signals. The numbers represent how many floating numbers are transmitted in communication.} 
        \label{fig_system_architecture}
\end{figure}

\subsection{Selective Communication}
Our proposed \textbf{selective communication algorithm} utilizes two rounds of communication, as shown in Fig. \ref{fig_system_architecture}. In the first round, each vehicle $s_i \in S_s$ uses its own lidar point cloud input $P^{s_i} \in \mathbb{R}^{N_p \times 3}$ and a lightweight lidar-based 3D object detector to predict the set of objects in its field-of-view, and sends the center locations of all detected objects to the eco-car. Let $D^{s_i} \in \mathbb{R}^{N_d^{s_i} \times N_v}$ represent vehicle $s_i$'s internal predicted object detection results, where $N_p$ is the number of lidar points, $N_d^{s_i}$ is the number of detected objects, and $N_v$ is the number of variables used to represent a detected object. In our implementation, $N_v = 7$ and the $j$th detected object ${D^{s_i}}_j$ can be represented by a vector $[[{D^{s_i}}_j]_x, [{D^{s_i}}_j]_y, [{D^{s_i}}_j]_z, [{D^{s_i}}_j]_l, [{D^{s_i}}_j]_w, [{D^{s_i}}_j]_h, [{D^{s_i}}_j]_a ]$, where $[[{D^{s_i}}_j]_x, [{D^{s_i}}_j]_y, [{D^{s_i}}_j]_z]$ represents the center position, $[[{D^{s_i}}_j]_l, [{D^{s_i}}_j]_w, [{D^{s_i}}_j]_h]$ represents the bounding box size, and $[{D^{s_i}}_j]_a$ represents the orientation. The $[D]c$ notation here represents the projection operator that accesses the $c$ column of the matrix $D$.
To minimize communication overhead in the first round, only a subset of this information, specifically only the center locations of detected objects projected to the 2D ground x-y plane, $[D^{s_i}]_{x,y} \in \mathbb{R}^{N_d^{s_i} \times 2}$, are transmitted to the ego-car.

%Next, each vehicle $s_i$ sends its object detection result information to the ego-car. Instead of sending whole detection result $D^{s_i}$, our algorithm only sends the detected objects' center positions projected to the 2D ground x-y plane, $[D^{s_i}]_{x,y} \in \mathbb{R}^{N_d^{s_i} \times 2}$,  to reduce the communication overhead in the first round of communication.

Once this information is received from all vehicles in $S_s$, the ego-car proceeds to estimate the information gain that would be provided by each prospective communicating vehicle. To do this, the ego-car compares its own set of 
detected object 2D center positions $[D^{ego}]_{x,y}$ based on it's own lidar input $P^{ego} \in \mathbb{R}^{N_p \times 3}$ to the set of  detected object 2D center positions $[D^{s_i}]_{x,y}$  received from each vehicle $s_i$ in $S_s$.  We define the \textbf{utility score} $U^{s_i}$ of vehicle $s_i$ as the number of objects that $s_i$ detects that are not detected by the ego-car. To simplify the calculation, we assume that $s_i$'s $j$th detected object ${D^{s_i}}_j$ is missed by the ego-car if and only if the 2D center positions of each detected object in $[D^{ego}]_{x,y}$ is at least $0.5$ meters away from $[{D^{s_i}}_j]_{x,y}$. Given this assumption, we define the \textbf{utility score} of $s_i$'s $j$th detected object as:
\begin{align}
    {U^{s_i}}_j = I\{d(
    [{D^{s_i}}_j]_{x,y}, 
    [{D^{ego}}_k]_{x,y},
    ) > 0.5 ~\forall ~k 
    \},
\end{align}
 where $I$ is an indicator function, $d$ is the Euclidean distance function, and $k$ is an index of the ego-car's detected object. The overall \textbf{utility score} $U^{s_i}$ of vehicle $s_i$ is then just the sum of the utility scores of all its detected objects:
\begin{align}
    U^{s_i} = \Sigma_{j=1}^{N_d^{s_i}} {U^{s_i}}_j
\end{align}
Once $U^{s_i}$ values have been computed for all $s_i \in S_S$, the vehicles associated with the highest $N_c$ utility scores are selected to form the communication scope $S_c$.
%Finally, we sort the agents in the selection scope $[s_1, s_2, ..., s_{N_s}]$ by their utility scores $[U^{s_1}, U^{s_2}, ..., U^{s_{N_s}}]$, and select the top $N_c$ with highest utility scores as the final communication scope $S_c = \{c_1, c_2, ..., c_{N_c}\}$.

In the second round of communication, each agent $c_i \in S_c$, upon being requested to, uses a point feature encoder to generate the encoded point feature observation vector $O^{c_i}$. These point features are then transmitted to the ego-car and used, together with a control neural network, to generate the final control signals.

\subsection{Perception and Control Neural Networks}
In this section, we first describe the model architecture of the perception network that is used as the 3D object detector in the first round of communication. Then we will summarize the model of the control network, including both the point feature encoder used by each vehicle in the second round of communication and the control network used in the ego-car to generate the final control signals.

\subsubsection{Perception Neural Network}
Instead of using state-of-the-art lidar-based 3D object detectors, we implement a lightweight detector in order not to introduce much computation overhead. The perception neural network we implemented is based on the ideas from Super Fast and Accurate 3D Object Detection (SFA3D) \cite{dung2020sfa3d} and Real-time Monocular 3D Detection (RTM3D) \cite{li2020rtm3d}.

\begin{figure*}[ht!]
\centering
\includegraphics[width=0.9\textwidth]{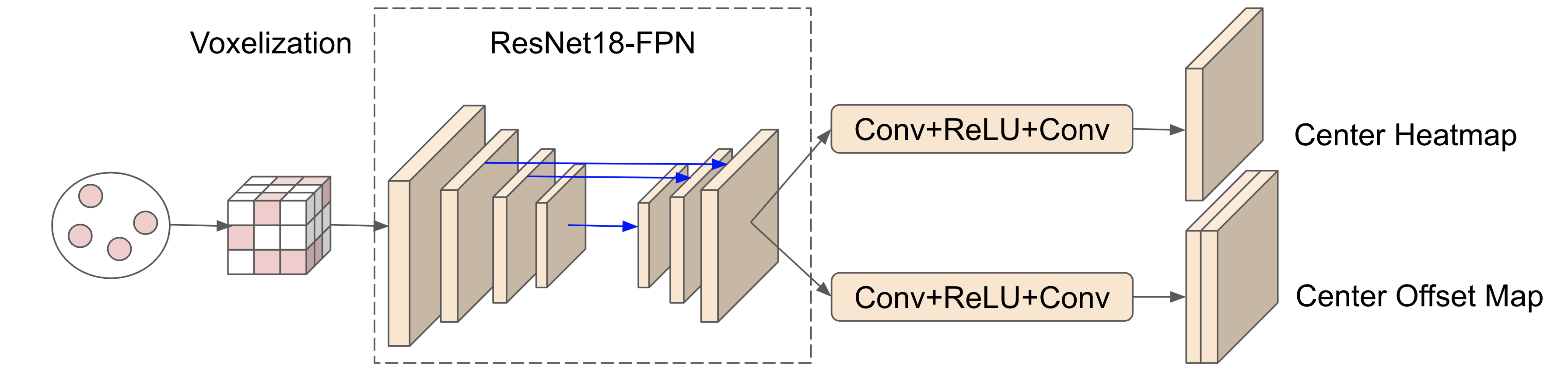}
\caption[]
        {Perception neural network, used as a lightweight 3D object detector.} 
%        We first voxelize the input point cloud to get a BEV feature map. Then the ResNet18 with Feature Pyramid Network is used to extract an intermediate feature map. Finally, two regression heads containing Conv and ReLU layers produce the center heatmap and the center offset map. We omit the illustrations of other outputs, such as object size and orientation, for simplicity because we do not use those attributes in the downstream process.
%        \vspace{-10pt}} 
        \label{fig_perception_neural_network}
\end{figure*}

Fig. \ref{fig_perception_neural_network} shows the computational pipeline of the implemented perception neural networks. The input point cloud is first voxelized to get a Birds Eye View (BEV) feature map. Then the ResNet18 with Feature Pyramid Network is used to extract an intermediate feature map. Finally, two regression heads containing Conv and ReLU layers are used to produce the center heatmap and the center offset map.

\begin{figure*}[h]
\centering
\includegraphics[width=0.9\textwidth]{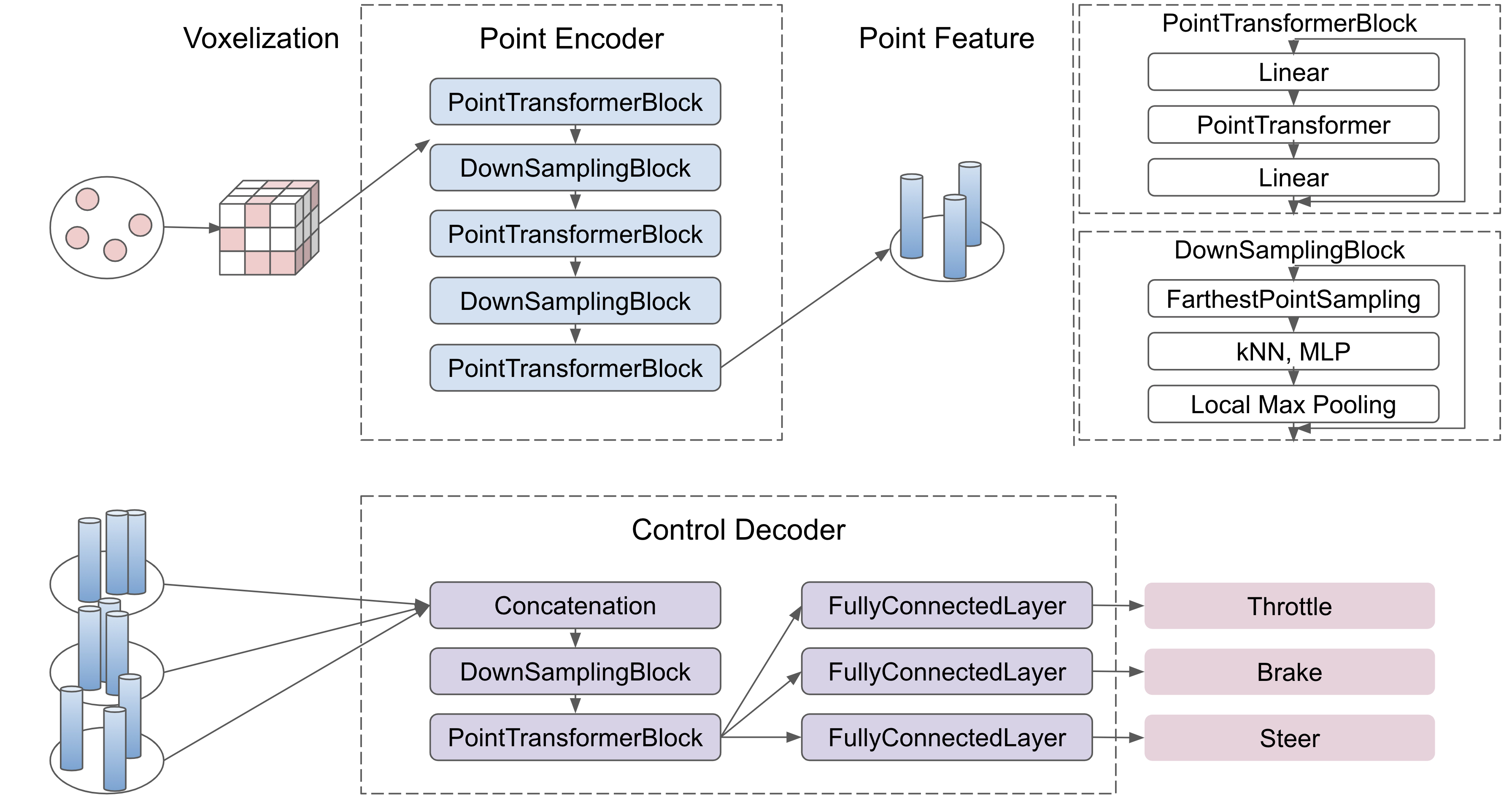}
\caption[]
        {Control neural network (same as that developed in \cite{cui2022coopernaut}). 
        %Each agent voxelizes the input point cloud and then uses the point transformer \cite{zhao2021point} block and downsampling block to get the point feature. All relevant agents' point features are sent to the ego-car, and then the ego-car uses the control decoder to aggregate those point features and produce the final control signals. For more details, please refer to the Appendix section.
        \vspace{-10pt}} 
        \label{fig_control_neural_network}
\end{figure*}

In more detail, each vehicle $s_i$'s raw lidar point cloud is down-sampled to $N_p = 2048$ points as $P^{s_i} \in \mathbb{R}^{N_p \times 3}$. The input BEV feature map $M^{s_i}_{in} \in \mathbb{R}^{X \times Y \times Z}$ is then generated by voxelizing the 3D space of the driving scene, whose size is $X_m = 140,  Y_m = 140, Z_m = 5$ in meters and centered by the agent $s_i$. The resolution of the voxelization is $dX = dY = dZ = 0.5$ meters, and that infers the shape of the BEV feature map $M^{s_i}_{in}$ is $X = \frac{X_m}{dX} = 280, Y = \frac{Y_m}{dY} = 280, Z= \frac{Z_m}{dz}  = 10$. Each voxel in the BEV map contains a binary value, indicating whether there exist at least $3$ lidar points in the voxel. This step can be described by 
\begin{align}
    M^{s_i}_{in} = V(P^{s_i}),
\end{align}
where $V$ is the voxelization function. \\

Next, the BEV feature map $M^{s_i}_{in}$ is used as the input to a Keypoint Feature Pyramid Network (KFPN) \cite{li2020rtm3d}. It uses ResNet18 \cite{he2015resnet} with Feature Pyramid Network (FPN) \cite{lin2017fpn} as the feature extraction backbone. This neural network generates an intermediate feature map as follows:
\begin{align}
    M^{s_i}_{inter} = ResNet18FPN(M^{s_i}_{in}),
\end{align}
where $M^{s_i}_{inter} \in \mathbb{R}^{\frac{X}{S} \times \frac{Y}{S} \times C}$ represents the intermediate feature map, $S = 4$ is the spatial scale factor, $C = 64$ is the feature channel size. Then the intermediate feature map $M^{s_i}_{inter}$ is used as the input of multiple regression head networks. Each regression head network contains $2$ Conv layers and a ReLU layer in between to generate an output map. Different output maps represent different important variables that can be used to construct the object detection result $D^{s_i} \in \mathbb{R}^{N_d^{s_i} \times N_v} $ as defined earlier. Two important regression heads and their output maps are:
\begin{align}
    M^{s_i}_{ch} &= R_{ch}(M^{s_i}_{inter}) \\
    M^{s_i}_{co} &= R_{co}(M^{s_i}_{inter}),
\end{align}
where $M^{s_i}_{ch} \in [0, 1]^{\frac{X}{S} \times \frac{Y}{S}}$ represents the predicted object center heatmap, $R_{ch}$ is the corresponding regression head neural network, $M^{s_i}_{co} \in \mathbb{R}^{\frac{X}{S} \times \frac{Y}{S} \times 2}$ represents the predicted center offset map, and $R_{co}$ is its regression head neural network.

Given a cell located at $(x_{ch},y_{ch})$ in the center heatmap $M^{s_i}_{ch}$ with the predicted object existence probability $M^{s_i}_{ch}[x_{ch},y_{ch}] > 0$, the corresponding center offset is
\begin{align}
    (x_{co}, y_{co}) = M^{s_i}_{co}(x_{ch}, y_{ch}).
\end{align}
The final predicted object 2D center position in the driving scene can be inferred based on the voxelization parameters as follows:
\begin{align}
    [D]x &= x_{ch} \times S \times dX - \frac{X_m}{2} + x_{co} \\
    [D]y &= y_{ch} \times S \times dY - \frac{Y_m}{2} + y_{co}.
\end{align}
Finally, we set a detection probability threshold value of $0.2$ and limit the number of objects detected by any given vehicle $s_i \in S_s$ to a maximum of $50$. The loss function we use to train the perception network is the focal loss \cite{lin2017focal} for the center heatmap $M^{s_i}_{ch}$ and the L1 regression loss for the center offset $M^{s_i}_{co}$. We omit the descriptions of other outputs, such as object size and orientation for simplicity, since those attributes are not used in the downstream process.

\subsubsection{Control Neural Network}

To enable a direct comparative basis for quantifying the performance gain due to use of our proposed selective communication model and our simplified lightweight perception network, we 
adopt the same control neural network that was originally developed in \cite{cui2022coopernaut}, and consider the results presented in this paper as the baseline. 
%in order to show that our model's performance gain is purely from our proposed selective communication algorithm and the simplified lightweight perception network, instead of tuning the control network architecture. 
Fig. \ref{fig_control_neural_network} shows the overall architectural diagram. Each vehicle voxelizes the input point cloud and then uses the point transformer \cite{zhao2021point} block and the downsampling block to get the point feature. The point features derived by each vehicle $c_i \in S_c$ are sent to the ego-car, after which the ego-car uses the control decoder to aggregate those point features and produce the final control signals. For details relevant to reproducing the results presented in the next section, please refer to the Appendix section. For full details of this control neural network, please refer to \cite{cui2022coopernaut}.

%In this section, we will just illustrate the architecture diagram in Figure \ref{fig_control_neural_network} and describe its key operations in the figure caption. For more details, please refer to the Appendix section.

\begin{figure*}[h!]
        \centering
        \begin{subfigure}[b]{0.32\textwidth}
            \centering 
            \includegraphics[width=\textwidth]{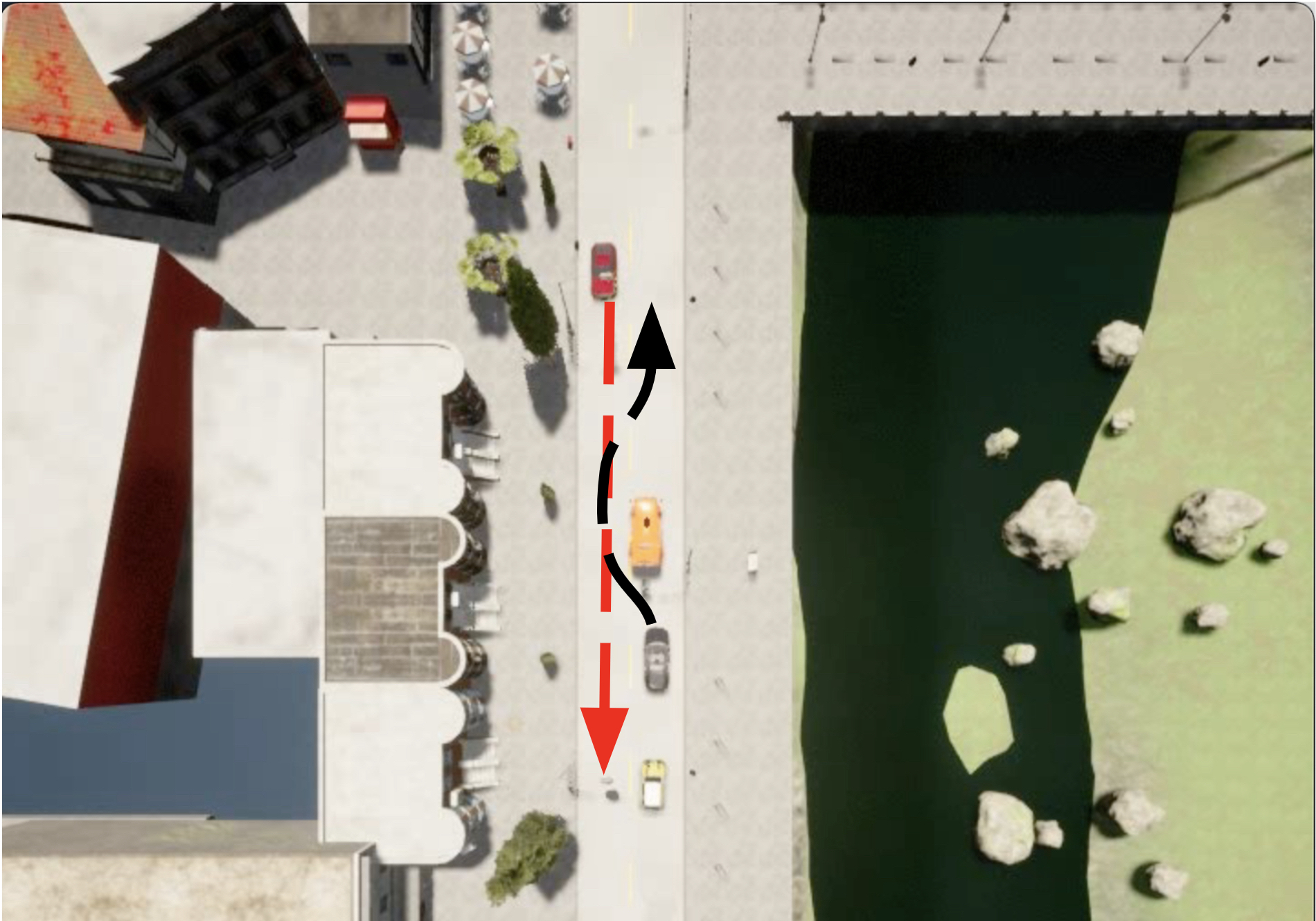}
            \caption[]%
            {{Overtaking}}    
            \label{fig_overtaking}
        \end{subfigure}
        \hfill
        \begin{subfigure}[b]{0.32\textwidth}  
            \centering 
            \includegraphics[width=\textwidth]{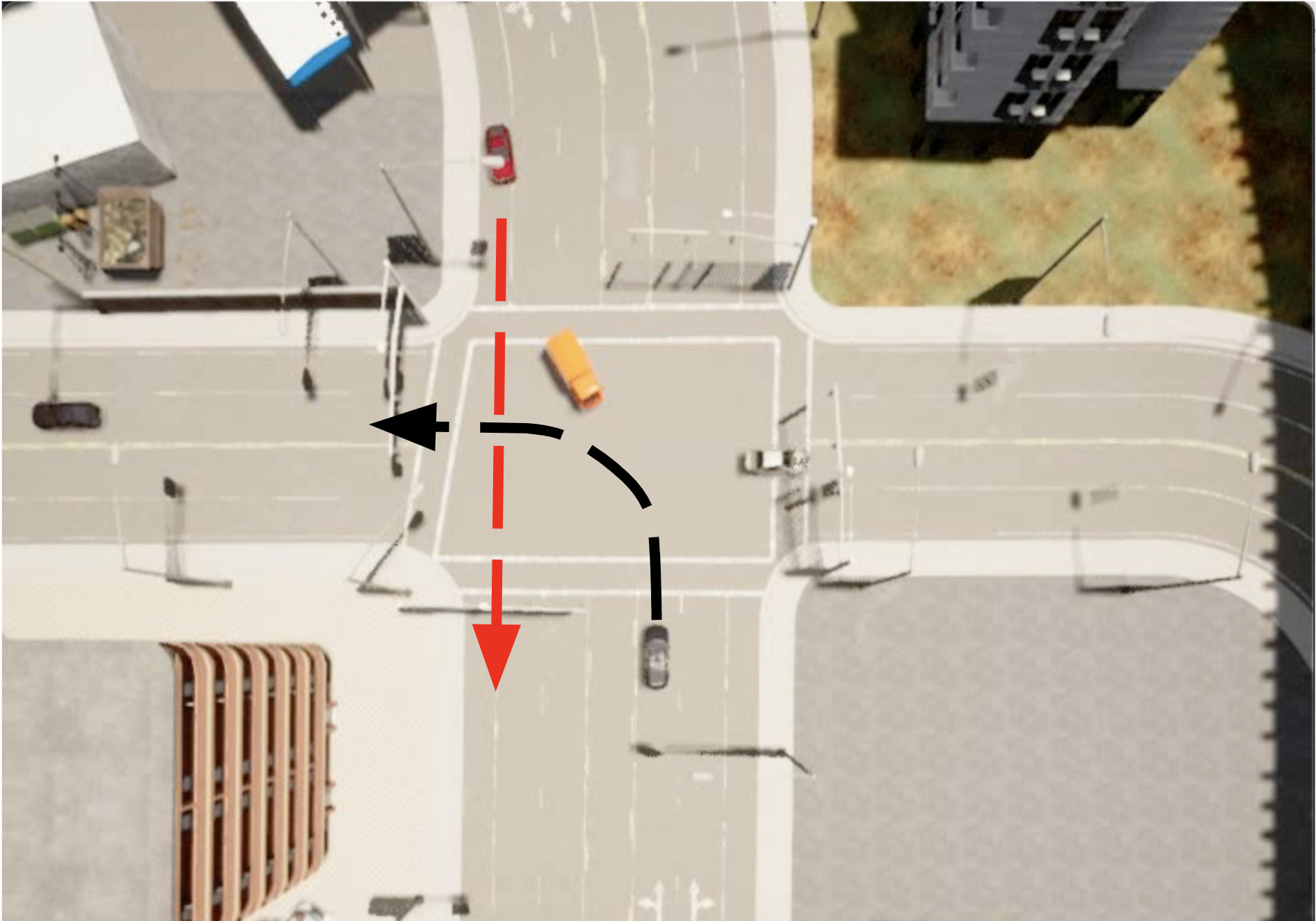}
            \caption[]%
            {{Left turn}}    
            \label{fig_left_turn}
        \end{subfigure}
        \hfill
        %\vskip\baselineskip
        \begin{subfigure}[b]{0.32\textwidth}
            \centering 
            \includegraphics[width=\textwidth]{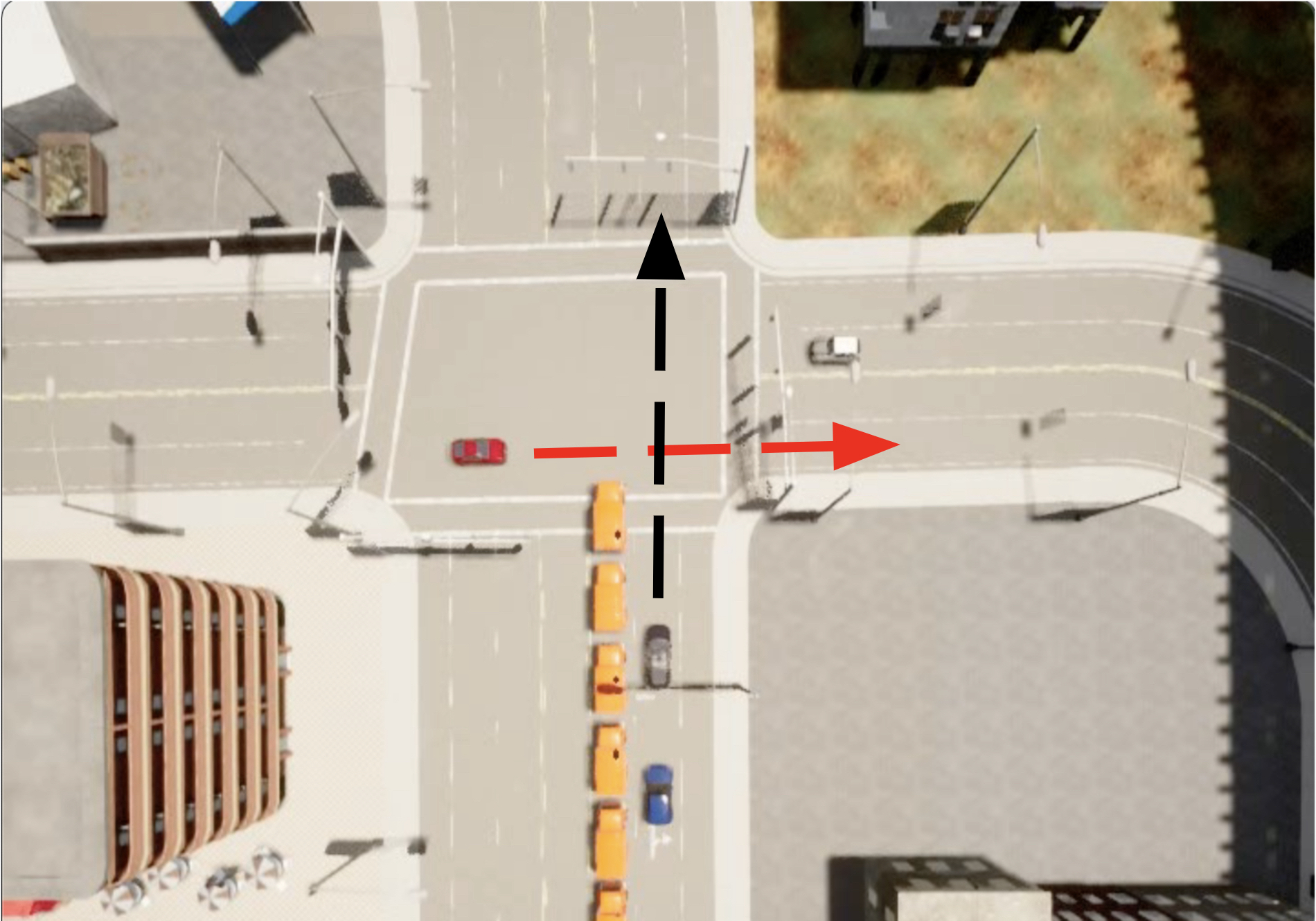}
            \caption[]%
            {{Red light violation}}    
            \label{fig_red_light_violation}
        \end{subfigure}
        \hfill
        \vspace{-5pt}
        \caption[]
        {
        Examples of the three categories of driving scenarios.} 
        \label{fig_three_scenarios}
    \end{figure*}

\begin{table*}[h!]
\small
\vspace{-5pt}
\caption{Evaluation results for $3$ types of driving scenarios, measuring average Success Rate (SR), Success rate weighted by Completion Time (SCT), Collision Rate (CR), and required bandwidth for single vehicles (single) and all vehicles (total).
%Quantitative evaluation results: evaluation in terms of average Success Rate (SR), Success rate weighted by Completion Time (SCT), and Collision Rate (CR) from multiple simulations with $30$ background traffic participants in the $3$ scenarios: Overtaking, Left Turn, and Red Light Violation. We also show the communication bandwidths for both single transmission of a vehicle and total transmission in all the vehicles.
\vspace{-15pt}}
\label{tab_evaluation_results}
\begin{center}
\begin{adjustbox}{width=1\textwidth}
\begin{tabular}{ l c cc ccc ccc ccc ccc}
  \hline
  \hline
  \multirow{2}{*}{Model} &
  \multirow{2}{*}{Selection Scope Size} &
  \multicolumn{2}{c}{Bandwidth (Mbps)} &
  \multicolumn{3}{c}{\textbf{Overtaking}} & \multicolumn{3}{c}{\textbf{Left Turn}} & \multicolumn{3}{c}{\textbf{Red Light Violation}} &\multicolumn{3}{c}{\textbf{Average}} \\
  \cline{3-16}
  & & single & total &
  SR $\uparrow$ & SCT $\uparrow$ & CR $\downarrow$ &
  SR $\uparrow$ & SCT $\uparrow$ & CR $\downarrow$ &
  SR $\uparrow$ & SCT $\uparrow$ & CR $\downarrow$ &
  SR $\uparrow$ & SCT $\uparrow$ & CR $\downarrow$
   \\
  \hline
  \hline
  Coopernaut~\cite{cui2022coopernaut} & 6 & 5.10 & 15.30 &
  90.5 & 88.4 & \textbf{4.5} &
  80.7 & \underline{76.2} & 18.1 &
  80.7 & 77.8 & 17.7 & 
  84.0 & 80.8 & 13.4 \\
  Selective Communication (Ours) & 6 & 5.13 & 15.48 &
  \textbf{92.6} & \underline{88.5} & 6.2 & 
  \underline{81.5} & \underline{76.2} & \textbf{16.0} &  
  \underline{82.7} & \underline{80.0} & \underline{14.8} &
  \underline{85.6} & \underline{81.6} & \underline{12.3} \\
  \hline
  Selective Communication (Ours) & 10 & 5.13 & 15.60 &
  \textbf{92.6} & \textbf{89.3} & \underline{4.9} & 
  \textbf{82.7} & \textbf{77.8} & \textbf{16.0} &  
  \textbf{84.0} & \textbf{81.4} & \textbf{12.3} &
  \textbf{86.4} & \textbf{82.3} & \textbf{11.1} \\
  \hline
  \vspace{-30pt}
\end{tabular}
\end{adjustbox}
\end{center}
\end{table*}

\section{Experimental Results}
\subsection{Settings}
To evaluate our approach to cooperative perception, we follow the same experimental design and evaluation settings utilized in \cite{cui2022coopernaut}. Three categories of pre-crash driving scenarios from the US National Highway Traffic Safety Administration \cite{najm2013description} are configured to construct safety-critical situations caused by visual occlusion as follows:
\subsubsection{Overtaking}
As shown in Fig. \ref{fig_overtaking}, a yellow truck remains static in front of the grey ego-car and prevents it from moving forward. The ego-car attempts to overtake by lane changing. However, another red car in the adjacent lane moving in the opposite direction is unable to be detected by the ego-car because the truck blocks the ego-car's line-of-sight. The arrows show that their planned trajectories will collide if cooperative perception is not applied.

\subsubsection{Left turn}
As shown in Fig. \ref{fig_left_turn}, the grey ego-car attempts to turn left in an intersection without a protected left-turn traffic light signal. However, a yellow truck remains static in the intersection waiting for its own left turn. The truck prevents the ego-car from fully perceiving another incoming red car moving in the opposite direction and makes it difficult for the ego-car to determine the right timing for proceeding with the left turn action.

\subsubsection{Red light violation}
As shown in Fig. \ref{fig_red_light_violation}, the grey ego-car attempts to cross an intersection. However, another red car in the horizontal lane ignores its red light signal and continues crossing. Moreover, a yellow truck blocks the ego-car's ability to detect the traffic rule-violating red car. 

In addition to the aforementioned three vehicles involved in the pre-crash scenarios, $30$ background vehicles are also introduced into the driving scenarios to produce simulations closer to real-world environments. Note that both the yellow truck and the background traffic participants can share their perception information with the ego-car, while the potential colliding red car can not. 
%Both baseline \cite{cui2022coopernaut} and our algorithm uses the ego-car's nearest $6$ vehicles as the selection scope, and select $3$ among them as the communication scope.

\subsection{Datasets}
We use the same training data from \cite{cui2022coopernaut}'s official website to train our lidar-based 3D object detection model. The training data contains $12$ driving sequences for each of the three scenarios. The data is generated by simulation using an oracle driving policy. This policy uses the ground-truth positions, orientations, sizes, and velocities of all vehicles to predict their future trajectories. Then the A* search algorithm is used to decide how the ego-car should act in every timestep in order to reach its goal without collision. 

For evaluation, we again follow the baseline paper \cite{cui2022coopernaut}. For each of the $3$ safety-critical scenarios, $27$ driving simulations are performed with $27$ different (but fixed) configurations of the attributes of the three main pre-crash relevant vehicles. We repeat each simulation configuration $3$ times with different random seeds to vary the starting and ending locations and times of background traffic participants. Both the algorithm of \cite{cui2022coopernaut} and our proposed variant, which differs only in the use of our two-round selective communication approach to determining $S_c$ instead of  \cite{cui2022coopernaut}'s original random selection approach, are evaluated with selection scope size $|S_s| = 6$ and communication scope size $|S_c| = 3$. Our proposed algorithm variant is also evaluated with $|S_s| = 10$ and $|S_c| = 3$ to examine the performance/communication tradeoff if the selecton pool is enlarged. 

\subsection{Metrics}
Each simulation ends with one of the three outcomes: success, collision, or stagnation. A simulation that is successful means the ego-car reaches its goal location within a limited time. If there is any collision with other agents or the environment, the simulation fails due to collision. If the ego-car is unable to reach the goal within a limited time, the simulation fails due to stagnation. We use success rates as the main evaluation metric. We also show the collision rates and another metric: success rates weighted by completion time (SCT) as follows:
\begin{align}
    SCT = I\{\text{success}\}T_{expert}/T_{model},
\end{align}
where $I$ is an indicator function, $T_{expert}$ is the completion time when simulating with the oracle expert driving policy, and $T_{model}$ is the completion time when using the proposed driving algorithm based on the trained model. Larger SCT indicates that the driving algorithm is more efficient without unnecessary braking.

\subsection{Quantitative Results}
\subsubsection{Performance}
%Both \cite{cui2022coopernaut} and our algorithm are tested with selection scope and communication scope values: $|S_s| = 6$ and $|S_c| = 3$. The only difference is that \cite{cui2022coopernaut} performs uniformly random selection while our proposed method sorts agents by our proposed utility scores.

Table \ref{tab_evaluation_results} shows the comparative performance results. 
%The first row contains the baseline \cite{cui2022coopernaut}, and our proposed method's results are shown in the second row. 
We can see that our proposed method (row 2) outperforms the baseline approach of \cite{cui2022coopernaut} (row 1) by $0.8\% \sim 2.1 \%$ for the 3 driving scenarios in terms of the success rates. The overall average success rates, success rates weighted by completion time, and collision rates are improved by $1.6\%$, $0.8\%$, and $1.1\%$ respectively. 
We performed a t-test over the binary outcomes produced by both approaches over all simulation runs (i.e., $1$ represents success and $0$ otherwise), yielding a p-value is $0.066$.
% showing that the performance gain is nearly statistically significant, based on the commonly used threshold value $0.05$.

Examining the performance difference that results from increasing the size of the selection scope to $|S_s| = 10$ (row 3 of Table \ref{tab_evaluation_results}), we see that 
%Moreover, we also explore the cases with a larger selection scope $|S_s| = 10$. The last row of Table \ref{tab_evaluation_results} shows that 
the overall average success rates, success rates weighted by completion time, and collision rates improved by  $0.8\%$, $0.7\%$, and $1.2\%$ respectively. 
%collision rate are further improved and outperform the baseline by $2.4\%$, $1.5\%$, and $2.3\%$ respectively. Performing the same t-test relative to the results of \cite{cui2022coopernaut} yields a p-value of $0.035$.
%showing that the improvement is statistically significant.

Overall these results confirm the performance advantage of intelligently selecting vehicles to cooperatively improve perception coverage. They also indicate that this performance advantage can be boosted by expanding the pool of nearby vehicles that are available to select from, suggesting that vehicles closest to the ego-car may also suffer similar perception occlusion problems and not always be the best choice for improving perception coverage. At the same time, these performance gains come at the cost of additional communication and it is important to quantify this tradeoff.

%Such improvement implies that intelligently selecting agents to cooperatively improve the perception coverage is critical to the final driving performance and safety. Additionally, agents near the ego-car may not always be the best choice because some of them may also suffer similar perception occlusion problems.

\subsubsection{Cost}
%The performance gain from our proposed selective communication algorithm does not come with no cost. 
The communication bandwidth requirements of each algorithm and configuration evaluated   is also reported in Table \ref{tab_evaluation_results}. In the baseline case \cite{cui2022coopernaut}, each of the selected $3$ vehicles transmits $128$ point features. Each point feature is a vector of size $(128 + 3)$ with 4-byte float numbers. Assuming the frame rate is $10$Hz, the communication bandwidth of each vehicle's transmission is $128 \times (128+3) \times 4 \times 10 = 670,720$ bytes per second, or roughly $5.10$ mega-bits per second (Mbps), as also reported in the baseline paper \cite{cui2022coopernaut}.

In our selective communication algorithm, each of the $6$ vehicles in $S_s$ first transmits (in the worst case) the 2D center positions of all detected objects to the ego-car. Assuming an upper bound of $50$ objects, this transmission uses $50 \times 2 \times 4 \times 10 = 4,000$ bytes per second, or roughly $0.03$ Mbps. The ego-car then requests the best $3$ vehicles to send the point features. Assuming each request can be accomplished by sending a single floating number, this step uses $1 \times 3 \times 4 \times 10 = 120$ bytes per second, which is 
roughly $0.0009$ Mbps. Combining with the bandwidth usage of \cite{cui2022coopernaut} for transmitting point features in communication round 2, the overall bandwidth requirement of each single vehicle is $0.03 + 0.0009 + 5.10 \sim= 5.13$ Mbps, an increase of just $0.6\%$ in communication overhead.

Since more vehicles are involved in communication when using the selective communication algorithm, we also calculate the total amount of data transmitted per second across  all vehicles. In \cite{cui2022coopernaut}, $3$ vehicles transmit point features to the ego-car. So the total data transmission is $5.10 \times 3 = 15.30$ Mbps. For the selected communication algorithm, each vehicle in $S_s$ first transmits the 2D center positions of its detected objects to the ego-vehicle, resulting in a total data transmission cost of $0.03 \times 6 = 0.18$ Mbps when $|S_s| = 6$. Requests are then sent from the ego-car to $|S_c| = 3$ vehicles, which costs $0.0009$ Mbps as 
before. Finally, the total cost incurred by these same $3$ vehicles sending their point features back to the ego-car is $15.30$ Mbps (the same as in \cite{cui2022coopernaut}). Hence, the the selective communication algorithm's total data transmission cost for $|S_s| = 6$ is $0.18 + 0.0009 + 15.30 \sim= 15.48$ Mbps, an increase of only $1.18\%$ over the baseline approach of \cite{cui2022coopernaut}.  When $|S_s| = 10$, the total data transmission cost increases by $0.78\%$ to  $0.03 \times 10 + 0.0009 + 15.30 = 15.60$ Mbps. 
%Our algorithm only introduces additional $1.18\%$ and $1.96\%$ of data transmission.

\subsection{Qualitative Results}
Please refer to the Appendix.

\section{Conclusion}
We propose a novel two-round selective communication algorithm for cooperative perception in end-to-end autonomous driving. In the first round, each vehicle in the ego-car's communication range performs lightweight 3D object detection and shares the 2D center positions of all detected objects with the ego-car. The ego-car then sorts those vehicles by the provided perception coverage gain and selects the best subset given the communication bandwidth constraints. In the second round of communication, each selected vehicle shares its encoded point transformer features with the ego-car and these features are aggregated to produce the control signals. 
Our algorithm improves on the success rates  achieved by the current state-of-the-art baseline in safety-critical driving scenario simulations, with only slight additional communication cost.
%Our algorithm outperforms the current state-of-the-art baseline in terms of success rates and other metrics of safety-critical driving scenario simulations.

For future work, we are investigating jointly training the perception and control neural networks with a multi-task learning approach.
%, which could further improve the performance of our algorithm.

% double-blind review
\section*{Acknowledgement}
The research reported in this paper was funded in part by the National Science Foundation under grant \#2038612, by the Google cloud research credit program, and by the CMU Robotics Institute.
% Google cloud research credit
%TODO

%%%%%%%%%%%%%%%%%%%%%%%%%%%%%%%%%%%%%%%%%%%%%%%%%%%%%%%%%%%%%%%%%%%%%%%%%%%%%%%%
%\clearpage
\balance
{\small
\bibliographystyle{IEEEtran}
\bibliography{egbib}
}

%%%%%%%%%%%%%%%%%%%%%%%%%%%%%%%%%%%%%%%%%%%%%%%%%%%%%%%%%%%%%%%%%%%%%%%%%%%%%%%%
\clearpage
\section*{Appendix}

\subsection{Control Neural Network (Details)}
We adopt the same control neural network architecture from the baseline \cite{cui2022coopernaut}. As shown in Figure \ref{fig_control_neural_network}, each selected vehicle $c_i$ in the communication scope $S_c$ takes its lidar point cloud $P^{c_i} \in \mathbb{R}^{N_p \times 3}$ as the input of a point cloud encoder neural network $E$, where $N_p = 2048$ is the number of lidar points. The point cloud encoder $E$ generates the encoded point cloud observation feature $O^{c_i}$ as follows:
\begin{align}
    O^{c_i} = E(P^{c_i}),
\end{align}
where $O^{c_i} \in \mathbb{R}^{\frac{N_p}{O_s} \times (O_f + 3)}$, $O_s = 16$ is the point down-sampling factor, $O_f = 128$ is the encoded feature size, and the constant $3$ represents that each point's feature vector of a size $O_f$ is appended by the point's 3D position vector. Overall the encoded observation output $O^{c_i}$ can be seen as a set of $\frac{N_p}{O_s}$ key points selected from the original input $N_p$ points, and each selected point contains a feature vector of a size $O_t = 128$ appended with its 3D position.

The point encoder $E$ contains two key components: the down-sampling block and the point transformer \cite{zhao2021point} block. The down-sampling block contains farthest point sampling, k nearest neighbor finding, multi-layer perceptron, and local max pooling. The main purpose of using the down-sampling block is to select important well-distributed key points from the input point cloud. The point transformer block contains the point transformer and linear layers. This block is used to aggregate important feature vectors from nearby points by an attention neural network \cite{vaswani2017attention}.

In the second round of communication, each agent $c_i$ in the communication $S_c$ sends its encoded point cloud observation feature $O^{c_i} \in \mathbb{R}^{\frac{N_p}{O_s} \times (O_f + 3)}$ to the ego-car. Then the ego-car concatenates its own encoded point cloud observation feature $O^{ego}$ and all the received features $\{O^{c_i} | c_i \in S_c \}$ as follows:
\begin{align}
    O^{aggr} = \odot(O^{ego}, O^{c_1}, O^{c_2}, ..., O^{c_{N_c}}) 
\end{align}
, where $O^{aggr} \in \mathbb{R}^{(\frac{N_p}{O_s}(N_c+1)) \times (O_f + 3)}$ is the aggregated point cloud observation feature, and $\odot$ is the concatenation operator. Then the aggregated feature $O^{aggr}$ is used as the input of the final control decoder $F$ to generate the final control signals as follows:
\begin{align}
    (throttle, brake, steer) = F(O^{aggr})
\end{align}
, where $throttle \in [0, 1]$, $brake \in [0, 1]$, and $steer \in [-1, 1]$ are the vehicle control signals to be applied to the ego-car, $F$ is the final control decoder neural network. The final control decoder also uses the down-sampling block and the point transformer block to model the interaction and correlation among point features from different agents. Finally, three regression heads implemented by fully-connected neural networks are used to predict the three control signals: $throttle, brake$, and  $steer$ as shown in Figure \ref{fig_control_neural_network}.

The training procedures for the control neural network contains two steps: behavior cloning and data aggregation training \cite{ross2015dagger}. We first run simulations with an oracle expert driving policy to collect the training data for behavior cloning. The oracle expert policy has access to all the obstacle locations and all agents' locations and velocities from the simulator. That information is used to predict other agents' future trajectories and find a collision-free space-time path by the A* search algorithm. The planned space-time path is then transformed to the control signals based on the vehicle dynamics as $(throttle_{oracle}, brake_{oracle}, steer_{oracle})$, which are used as the output part of the training data. The lidar point clouds from the ego-car and the selected vehicles in the communication scope $(P^{ego}, P^{c_1}, P^{c_2}, ..., P^{c_{N_c}})$ are used as the input part of the training data.

During the training of behavior cloning, we feed the input part of the training data $(P^{ego}, P^{c_1}, P^{c_2}, ..., P^{c_{N_c}})$ to the control neural network and get the network's output control signals $(throttle_{nn}, brake_{nn}, steer_{nn})$. And we calculate the loss $L$ by the sum of the L1 losses between the control network's output signals and the oracle expert driving policy's control signals as follows:
\begin{align}
    L = &|throttle_{nn} - throttle_{oracle}| ~ + \\
        &|brake_{nn} - brake_{oracle}| ~ + \\
        &|steer_{nn} - steer_{oracle}|
\end{align}

Once we have the trained model from behavior cloning, we use it as the initialization model for the data aggregation training. In data aggregation training, we iteratively perform interleaving data collection and model training procedures. But the simulation steps in data collection are not always using the oracle expert policy's driving control signals. Instead, there exists an exponentially decayed probability to decide whether to use the control signals from the oracle expert policy or the model being trained. The main motivation of this data aggregation training procedure is to explore more possible scenarios that the current model may experience but the expert policy has not seen before. And the expert policy can provide better control signals for those scenarios as the training data in the next model training iteration.

\subsection{Qualitative Results}
The driving video samples of the simulation results using our trained model can be found in the url ( 
\url{https://youtube.com/playlist?list=PLJDt5JmFddFemsN-8du-TlUXvD6jAznpc}
%\url{https://drive.google.com/drive/folders/1qymBxmgFOoxN3JAxCPfxMafYWDsmi6Oa}
). In general, we can see that the ego-car controlled by our proposed algorithm is aware of other occluded traffic participants and able to make correct driving decisions to avoid collisions.

%%%%%%%%%%%%%%%%%%%%%%%%%%%%%%%%%%%%%%%%%%%%%%%%%%%%%%%%%%%%%%%%%%%%%%%%%%%%%%%%

\end{document}